\documentclass{article}

\usepackage{arxiv}

\usepackage[utf8]{inputenc} % allow utf-8 input
\usepackage[T1]{fontenc}    % use 8-bit T1 fonts
\usepackage{xcolor}
\usepackage{hyperref}       % hyperlinks
\hypersetup{%
  colorlinks=false,
  urlbordercolor=white
}
\usepackage{xurl}
\usepackage{booktabs}       % professional-quality tables
\usepackage{amsfonts}       % blackboard math symbols
\usepackage{nicefrac}       % compact symbols for 1/2, etc.
\usepackage{microtype}      % microtypography
\usepackage{lipsum}
\usepackage{graphicx}

\usepackage[caption=false]{subfig}

\usepackage{amsmath,amssymb,amsfonts,amsbsy,amsthm,booktabs,epsfig,graphicx,rotating}

\usepackage{algorithm}
\usepackage{algpseudocode}
\algnewcommand\algorithmicforeach{\textbf{for each}}
\algdef{S}[FOR]{ForEach}[1]{\algorithmicforeach\ #1\ \algorithmicdo}
\usepackage{todonotes}
\usepackage{url}

\graphicspath{ {./images/} }

\title{Visual Trajectory Prediction of Vessels for Inland Navigation}

\usepackage{authblk}

\newbox{\orcid}\sbox{\orcid}{\includegraphics[scale=0.06]{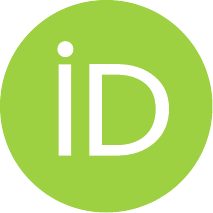}} 
\author[1]{%
	\href{https://orcid.org/0000-0002-3851-5507}{\usebox{\orcid}\hspace{1mm}Alexander Puzicha\thanks{\texttt{CONTACT Alexander Puzicha. Email: alexander.puzicha@cs.tu-dortmund.de}}}%
}
\author[1]{%
	\href{https://orcid.org/0000-0003-2525-2477}{\usebox{\orcid}\hspace{1mm}Konstantin Wüstefeld}%
}
\author[2]{%
	Kathrin Wilms%
}
\author[2]{%
	\href{https://orcid.org/0000-0002-2530-8197}{\usebox{\orcid}\hspace{1mm}Frank Weichert}%
}
\affil[1]{Chair of Modeling and Simulation, Computer Science, TU Dortmund University, Germany}
\affil[2]{Chair of Computer Graphics, Computer Science, TU Dortmund University, Germany}

\begin{document}
\maketitle
\begin{abstract}
The future of inland navigation increasingly relies on autonomous systems and remote operations, emphasizing the need for accurate vessel trajectory prediction. This study addresses the challenges of video-based vessel tracking and prediction by integrating advanced object detection methods, Kalman filters, and spline-based interpolation. However, existing detection systems often misclassify objects in inland waterways due to complex surroundings. A comparative evaluation of tracking algorithms, including BoT-SORT, Deep OC-SORT, and ByeTrack, highlights the robustness of the Kalman filter in providing smoothed trajectories. Experimental results from diverse scenarios demonstrate improved accuracy in predicting vessel movements, which is essential for collision avoidance and situational awareness. The findings underline the necessity of customized datasets and models for inland navigation. Future work will expand the datasets and incorporate vessel classification to refine predictions, supporting both autonomous systems and human operators in complex environments.
\end{abstract}

% keywords can be removed

\section{Introduction}
\label{sec:Introduction}
The future navigation of inland vessels increasingly relies on remote operators and autonomous systems. In such scenarios, operators have access to video streams but lack critical spatial and environmental depth information, which complicates decision-making. Accurate trajectory prediction of vessels is essential for enhancing situational awareness, ensuring safe maneuvering, and supporting autonomous navigation.

In this study, we address these challenges by integrating advanced object detection and tracking methodologies with predictive modeling tailored to the dynamic behavior of inland vessels. Our focus is on analyzing and predicting vessel trajectories using computer vision techniques combined with a Kalman filter and spline-based interpolation methods. These tools aim to overcome the limitations of conventional detection systems, such as inaccuracies in Automatic Identification System (AIS) data due to misconfigured sensors, and to provide robust tracking and prediction capabilities.

Based on modern object detection and advanced tracking algorithms, we investigate the accuracy of multi-object tracking and prediction for inland and sea-going vessels on waterways.

This work contributes to the growing body of research on inland waterway navigation by addressing gaps in vessel detection and trajectory prediction. The results are expected to enhance the safety and efficiency of human-operated and autonomous vessel navigation.

\section{Literature review}
\label{sec:literature}
As recent studies have shown~\cite{Henn.2018, Martelli.2021, Theotokatos.2023, Xu.2023}, camera-based systems are mandatory for the future development of remote-controlled and fully autonomous vessels on oceans and inland waterways. Their task mainly focuses on surveillance, such as the detection of ships and other objects, tracking, and classification to replace or support captains in collision avoidance.
Vessel detection and tracking consist of per-frame vessel detection, object-based classification to distinguish different instances, and vessel tracking over multiple frames. There are general tracking approaches~\cite{tracking2019} to solving the tasks without special adaptations to the problems of shipping and specific approaches. In the field of shipping, a distinction is also made between ocean shipping and inland shipping~\cite{Henn.2018}.\\
In the area of object detection, various approaches can be identified with regard to the training data and label information, which use different degrees of label information: supervised learning~\cite{liu2020deep}, semi-supervised learning~\cite{wang2023semi}, few-shot learning~\cite{kohler2023few}, weakly supervised learning~\cite{shao2022deep}, and unsupervised learning~\cite{zhao2019object}. The available model architectures are divided into two main groups:
There are two-stage and one-stage object detectors, relying on CNN- or transformer-based object detection models~\cite{zaidi2022survey}. %Two-stage object detection models use region proposals~\cite{girshick2014rich}. 
%
%Most approaches determine maps of relevant features from an image as a first step using a pre-trained network from the ResNet family~\cite{ResNet,Liu2020}. After that, regions with objects of interest are proposed from the generated feature maps, and the object class is predicted based on the confidence of the model in terms of how plausible the classification is.
%
An example of a one-stage object detector is the YOLO (You Only Look Once) architecture~\cite{diwan2023object}.
%YOLOv8 is a recent derivative of the initial YOLO network. 
This family of networks~\cite{YOLO_family} follows another paradigm of proposing bounding boxes. While in two-stage approaches, the proposed regions are the result of the classification of several fixed anchor regions~\cite{FasterRCNN}, each cell of a grid that is applied over the feature maps in YOLO proposes bounding boxes as a regression task. By refraining from multi-stage analysis, such networks reach higher computational efficiency~\cite{YOLO_family}.
Transformer-based object detection is based on the fundamental architecture of the Detection Transformer (DETR)~\cite{carion2020end}. DETR networks rely on an encoder-decoder transformer~\cite{AttentionIsAll} architecture and a bipartite graph matching loss to simplify bounding box prediction, making the use of fixed anchors or cell regions, as used in other approaches, unnecessary. An advantage of that is the avoidance of implicit biases emerging from the specification of possible sizes, positions, or resolutions when determining anchor boxes while claiming results on a par with Faster R-CNN~\cite{DETR}.
RT-DETR~\cite{RT-DETR} is the successor of the basic DETR, maintaining its detection performance in most cases while claiming to beat YOLO networks in real-time tasks. A known drawback is the inferior performance in cases with small objects~\cite{DETR}, which could lead to limitations when boats are far away from the camera.
%RT-DETRv2 introduces a new training strategy improving the overall accuracy of RT-DETR without a loss in computational efficiency~\cite{RT-DETRv2}.
%
A common two-stage detector is Faster R-CNN (Faster Regions with CNN
features)~\cite{FasterRCNN}.
Faster R-CNN promises precise object detection through its division into specialized sub-networks, but despite being the fastest within the R-CNN family, it compromises on reduced speed compared to YOLO and RT-DETR~\cite{YOLO_family, RT-DETR}.\\

Multiple Object Tracking (MOT) has developed in a similar way to detection because the standard approach for MOT algorithms is to track by detection. For a detailed overview, refer to the corresponding overview papers~\cite{ciaparrone2020deep, hassan2024multi}. However, the development of techniques specifically for ships has lagged far behind other objects, such as vehicle detection and tracking~\cite{wang2022review} or traffic sign recognition~\cite{zhu2022traffic}. This is despite the fact that vessel detection involves challenges comparable to generic object recognition, such as lighting, occlusions, multiple categories, and variations within a class.
Furthermore, developments in this area tend not to follow a systematic pattern. Accordingly, there are only a few overview publications dealing with the use of deep learning methods to determine ship movements, e.g., \cite{er2023ship, bo2021ship, sasikala2019ship} or on specifically available data sets~\cite{zhang2024development}. State-of-the-art is primarily characterized by individual solutions.
Common examples of tracking algorithms that we chose to compare in our evaluation are Deep OC-SORT~\cite{Deep_OC-SORT}, 
BoT-SORT~\cite{aharon2022bot}, and ByeTrack~\cite{zhang2022bytetrack}.\\
%, whose advantages are detailed in the following paragraphs:
Deep OC-SORT~\cite{Deep_OC-SORT} focuses especially on improving results for similar objects occluding each other by combining a re-identification network~\cite{FastReID}.
% with a motion compensation of camera movements followed by a Kalman filter to estimate the current object position from the candidate track. The information from BoT-SORT paths is weighted adaptively and used in a solver to decide on the final assignment of new single-frame detections to tracks. 
While many tracking methods first discard one-image detections that have low confidence, ByteTrack~\cite{zhang2022bytetrack} also includes those with low confidence to achieve a higher recall through more possibilities of association between entries.
%High-confidence detections are matched directly based on Kalman filter predictions and intersection over union~(IoU) values. Low-confidence detections are used to bridge gaps between fragmentary high-confidence occurrences in a track~\cite{ByeTrack}. Due to this structure, ByeTrack promises an improvement, especially in applications with small objects or image disturbances.
BoT-SORT~\cite{aharon2022bot} shows similarities to the aforementioned approaches: it uses a differentiation between high- and low-confidence detections like ByteTrack and shows a two-path structure as in Deep OC-SORT.
%, including the use of a camera motion estimation before the Kalman filter. BoT-SORT is rated superior to ByeTrack and Deep OC Sort on pedestrian tracking data sets by its creators~\cite{BoT-SORT}.
%

The leading companies (e.g., SeaAI) for ocean vessel detection\footnote{Robosys and SEA.AI create machine vision-based maritime autonomous surface collision avoidance system, \url{https://www.maritimeindustries.org/news/robosys-and-seaai-create-machine-vision-based-maritime-autonomous- surface-collision-avoidance-system} [last referenced: 01-05-2025]} use a two-stage approach to detect and afterward classify floating objects.
Their system reveals weaknesses for inland waterways with a high rate of false positive detections because it is trained to detect every floating object on empty seas and not to extract vessels out of obstructed areas. Remote sailing companies for inland waterways (e.g., Seafar\footnote{Seafar, \url{https://seafar.eu} [last referenced: 01-05-2025]}) use YOLO for vessel detection.
%, which we will prove to have weaknesses for inland vessel detection as well \todo[color=orange]{nur, wenn die Auswertung das auch hergibt}.
Previous research on ship detection and tracking of ships, e.g., \cite{jie2021ship}, used the improved YOLOv3~\cite{farhadi2018yolov3}, Deep Simple Online and Real-time Tracking (Deep SORT~\cite{Wojke2017}) tracking algorithm.
% similar to the industry approaches\todo[color=orange]{Cite fehlt.Wenn nicht belegbar, Aussage entfernen. Nicht durch Zitat belegbar.}.
Other existing approaches to machine learning-based ship detection and tracking using satellite images for maritime surveillance~\cite{wang2021machine} differ significantly from the proposed methodology because of the vessel-to-vessel perspective. In inland navigation scenarios, relatively long vessels appear close to the cameras; thus, often, only parts of vessels are observed. In satellite images, almost always the entire ship is depicted, and its top-view shape is similar for most vessels. This does not hold for its side perspective.

\section{Methodology}
\begin{figure}[t]
	\includegraphics[width=\linewidth]{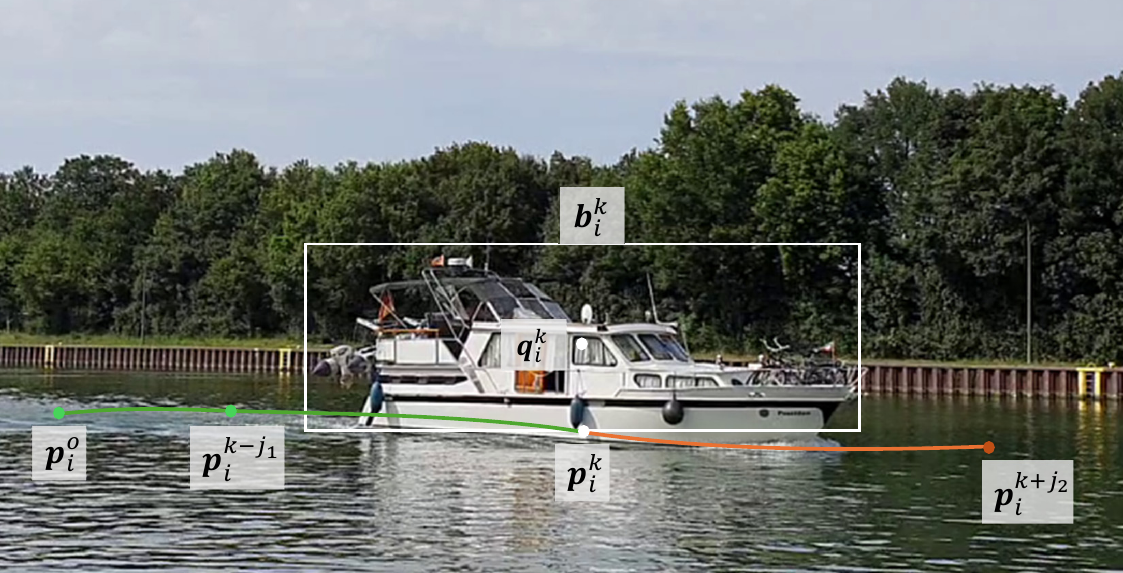}	
	\caption{Visualization of the parameters used for detection, tracking, and prediction on the example of a ship moving in the forward direction. The bounding box $\mathbf{b}_i^k$, the center of the box $\mathbf{q}_i^k$, the current anchor point $\mathbf{p}_i^k$ used for tracking, the trajectory of past frames (green) and the predicted trajectory (orange) with corresponding sample times for the current frame with the index $k$ and the object with the index $i$. The values $j_1>0$ and $j_2 > 0$ are used to illustrate the points in time.}
\label{fig:params_img}
\end{figure}
The following section details our methods for predicting the trajectory of vessels. In Section~\ref{subsec:semantic_tracking}, we describe our approach for the detection and tracking of vessels, and in Section~\ref{sec:Trajectory_prediction}, we outline methods for prediction. As an overview for the following subsections, Figure~\ref{fig:params_img} illustrates the parameters used in detection, tracking, and prediction with an example image. For each frame $k$, the object $i$ is characterized by a bounding box $\mathbf{b}_i^k$. An anchor point $\mathbf{p}_i^k$ is determined for each bounding box. Anchor points $\mathbf{p}_i^0$ to $\mathbf{p}_i^k$ are used for tracking (green trajectory). The anchor points $\mathbf{p}_i^k$ to $\mathbf{p}_i^{k+j_2}$ are predicted using a spline based on the previous values (orange trajectory).

\subsection{Semantic tracking}
\label{subsec:semantic_tracking}
Two preparatory steps are required to forecast the movement of objects: The detection of an arbitrary number of objects on the individual frames of a video and the tracking, i.e., the assignment of detected objects on successive frames to an object that moves over time. Although the changes in the position of a vessel are generally small in relation to common frame rates, external influence on the camera can cause trembling movements of the entire image, especially if the recording camera is positioned on a boat. This and other types of artifacts should, therefore, be considered in the detection process.

\subsubsection{Multi-Object Detection}
We define the image detection function as 
\begin{align}
\begin{split}
\mathcal{B}(\mathbf{I}_k) &= \left\{ \left( \mathbf{b}_i^k, c_i^k, e_i^k, \right) | i \in \{1, ..., n_k\} \right\}\\
&= \left\{ \mathcal{D}_i | i \in \{1, ..., n_k\} \right\}\\
\end{split}
\end{align}
that specifies the bounding box $\mathbf{b}$, class $c$, and confidence $e$ in the detection correctness for each of the $n_k$ detections on the $k$-th frame of the input video.

\begin{figure}[t]
\centering
\subfloat[]{\includegraphics[width=.33\textwidth]{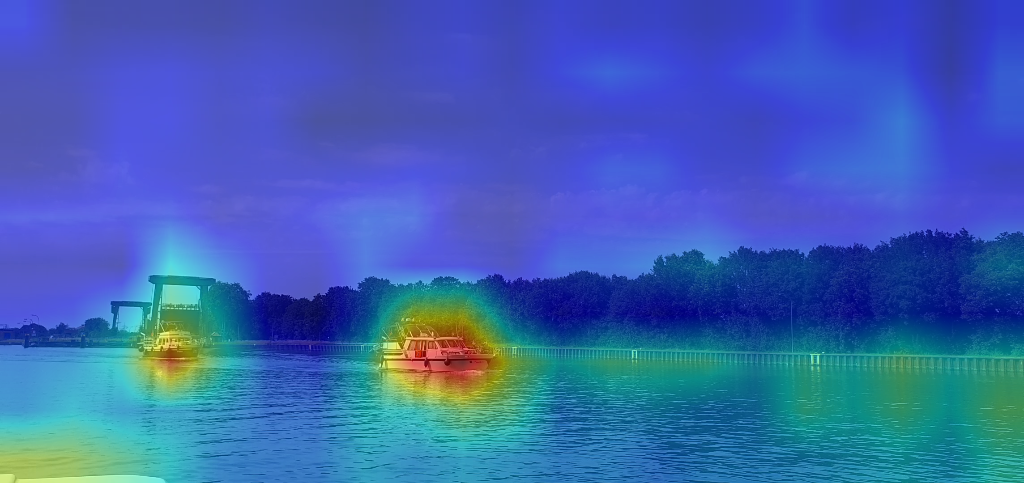}}\hfill
\subfloat[]{\includegraphics[width=.33\textwidth]{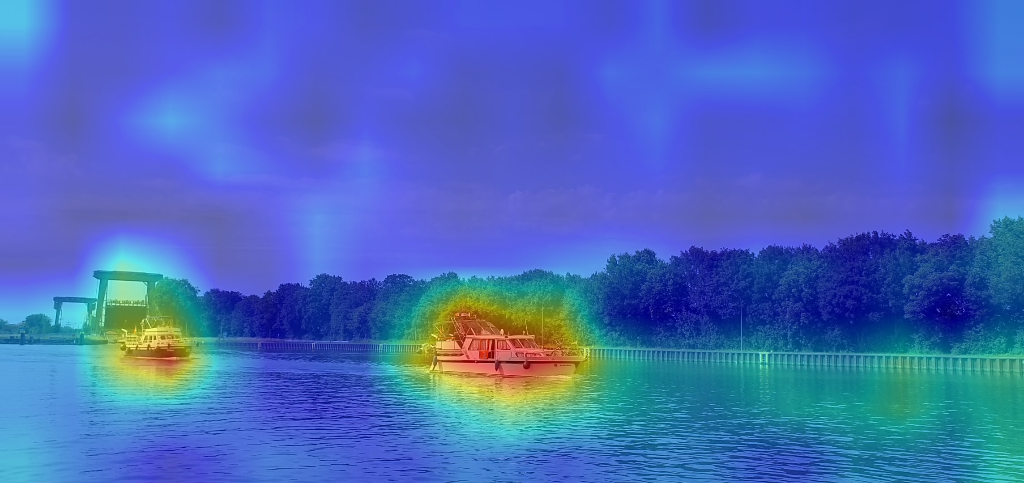}}\hfill
\subfloat[]{\includegraphics[width=.33\textwidth]{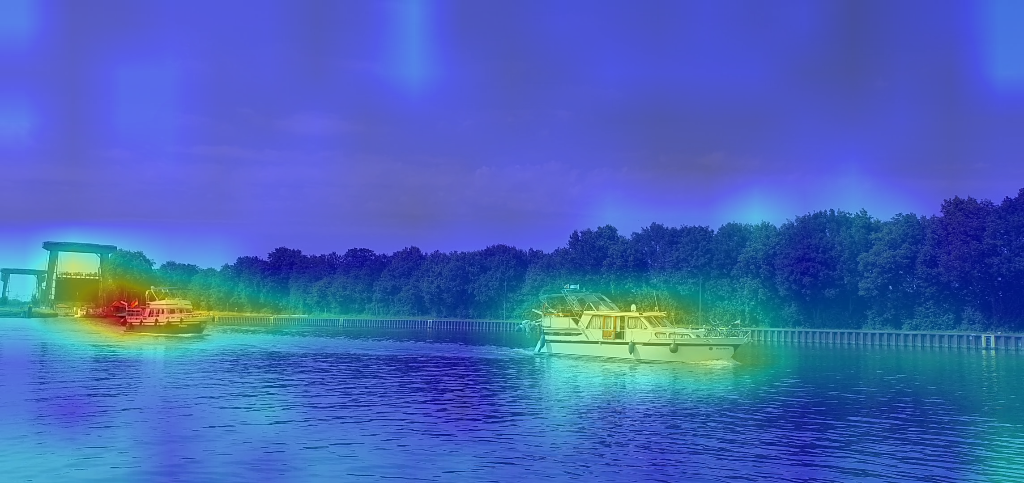}}
\caption{Relevance visualization of image areas for the assignment of objects to the boat class using a Faster R-CNN detector trained on the Microsoft Common Objects in Context~(COCO) data set~\cite{COCO}. Based on the result values of the Score-CAM~\cite{ScoreCAM} method, the overlay of the respective original image with red indicates a higher relevance, with yellow a medium relevance and with blue a lower relevance. The three images show consecutive frames from the same video where vessels are moving in the forward direction. Note the slight camera movement, which complicates tracking due to a changed perspective.}
\label{fig:cam1}
\end{figure}

In this context, the bounding box $\mathbf{b}_i^k=\left(\mathbf{q}_i^k, w_{k,i}, h_{k,i}\right)$ marks the center $\mathbf{q}_i^k$ and the expansion $\left(w_i^k, h_i^k\right)$ of the minimal box containing the region of the $i$-th detection in the $k$-th frame.
While the goal is to describe each object with exactly one detection, not every object is necessarily assigned one box. Depending on the specific detection method, there may be different boxes assigned or none at all. In the special case of significantly overlapping bounding boxes, the corresponding detections can be combined into one, e.g., by averaging the values from each of the detections to merge. In Fig.~\ref{fig:cam1}, class activation mapping (CAM) techniques are used to explain CNN detection~\cite{he2022survey}.

\subsubsection{Multi-Object Tracking}
\label{subsec:Multi-Object Tracking}
An algorithm for multi-object tracking in a video $\mathbf{V}=(\mathbf{I}_0, ..., \mathbf{I}_J)$ can be abstracted to a general form described in Algorithm~\ref{alg:tracking_abstract}. As a simplification, we assume that the tracked entities, in this case ships, do not split or merge.
A concrete multi-tracking algorithm implements and possibly extends these abstract steps giving specialized functions $h$ and $g$: $h(\mathcal{D},\mathcal{T})$ determines the plausibility that a detection $\mathcal{D}$ belongs to track $\mathcal{T}$ and $g(\mathcal{T})$ is a deletion criterion that is true if $\mathcal{T}$ is seen as an invalid track.

\begin{algorithm}
\renewcommand{\algorithmicrequire}{\textbf{Input:}}
\newcommand{\BREAK}{\STATE \algorithmicbreak}
		
\caption{Abstract algorithm for multi-object tracking methods.}\label{alg:tracking_abstract}
\begin{algorithmic}
\Require Video $V=(\mathbf{I}_0, ..., \mathbf{I}_J)$, per-frame detection $\mathcal{B}$, merge filter $h$, track deletion filter $g$
\State $\mathcal{T}_1=\mathcal{B}(\mathbf{I}_1)$
\ForEach {$j \in \{2,...,J\}$}
	\State $\mathcal{T}'_j = \varnothing$	
	\ForEach{$D \in \mathcal{B}(\mathbf{I}_j)$}
		\State flag $\gets$ 0
		\ForEach{$\mathcal{T} \in \mathcal{T}_{j-1}$}
			\If{$h(D, \mathcal{T}) > h(\mathcal{D}', \mathcal{T})$ for all $\mathcal{D}' \in \mathcal{B}(\mathbf{I}_j)$}
				\State update $\mathcal{T}$ by merging it with $D$
				\State flag $\gets$ 1			
			\EndIf
		\EndFor
		\If {flag $= 0$}
			\State $\mathcal{T}'_j \gets \mathcal{T}'_j \cup D$
		\EndIf
	\EndFor
	\State $\mathcal{T}_j \gets \mathcal{T}'_j \setminus \{g(\mathcal{T}) | \mathcal{T} \in \mathcal{T}'_j\}$
\EndFor
\end{algorithmic}
\end{algorithm}

To obtain more meaningful positions for tracking and visualization, we calculate the point $\mathbf{p}_i^k=\left(\mathbf{q}_i^k-\begin{pmatrix}0\\0.5 \cdot h_i^k\end{pmatrix}\right)$ from the coordinates of a box $\mathbf{b}_i^k$, which represents the center of the contact line between the vessel and the water.

\subsection{Trajectory prediction}
\label{sec:Trajectory_prediction}
As explained in the introduction (see Section~\ref{sec:Introduction}), external or autonomous navigation requires the identification of critical situations (e.g., collisions) at an early stage. Therefore, the prediction of vessel movements is addressed in the present work. This implies, based on the trajectory (see Section~\ref{subsec:Multi-Object Tracking}) $\mathcal{T}_i=\{\mathbf{p}_i^0, \ldots, \mathbf{p}_i^j, \ldots, \mathbf{p}_i^k\}$, $k$ the current frame, the prospective positions $\mathbf{p}_i^j$, for $j=k+h$, with $h>0$, are to be predicted. Since the movement trajectory of inland waterway vessels tends to have a pronounced smoothness, cubic Hermite splines are used as the interpolation method to determine the trajectories. These are suitable due to the availability of efficient algorithms for calculation and analysis~\cite{franke1991scattered}. The function within the relevant intervals $\mathbf{p}_i^l$ and $\mathbf{p}_i^{r}$ is interpolated by third-degree polynomials. The general cubic spline is given in parametric form according to
\begin{equation}
\mathbf{f}(t)=\mathbf{a}t^3 + \mathbf{b}t^2 + \mathbf{c}t + \mathbf{d}\textrm{,}
\end{equation}
with $t\in [0, 1] $~\cite{khan2005survey}. Using the constraints for Hermite splines, the following representation for determining the coefficients is obtained~\cite{wagner2010path}:
\begin{equation}
\mathbf{f}(t) = \left[ t^3 ~ t^2 ~ t ~ 1 \right]
\left[ \begin{array}{cccc}
2  & -2 & 1 & 1\\
-3 & 3 & -2 & -1 \\
0 & 0 & 0 & 0 \\
1 & 0 & 0 & 0\\
\end{array} \right]
\left[ \begin{array}{l}
\mathbf{p}_i^l \\ \mathbf{p}_i^{r}\\ \mathbf{t}_{\mathbf{p}_i^l} \\ \mathbf{t}_{\mathbf{p}_i^{r}}\\ 
\end{array} \right]\textrm{.}
\end{equation}
Here, $\mathbf{t}_{\mathbf{p}_i^l}$ and $\mathbf{t}_{\mathbf{p}_i^{r}}$ are the direction vectors at the points $\mathbf{p}_i^l$ and $\mathbf{p}_i^{r}$, which result from the adjacent points. The sampling of the points to be considered is carried out according to the Chebyshev condition~\cite{occorsio2021filtered}. Since ships in inland waterways normally have a fairly smooth trajectory, $l=0$ and $r=k$ apply. If this is not the case, the angle between $\mathbf{t}_{\mathbf{p}_i^l} $ and $\mathbf{t}_{\mathbf{p}_i^r}$ is greater than a specified error $\varepsilon$, the trajectory is split into a separate segment for which the interpolation is then carried out. The spline is used to regularize the determined bounding boxes in order to combine smaller variations in the detection. In addition, the spline is used to predict future movements of the ships. If the entire trajectory has been separated into subsegments, the last subsegment is used for prediction. One research question here is the quality of a forecast in the forecast period $h$ for inland waterway vessels and for sea-going vessels.\\

In addition to the spline embedding, a Kalman filter has been integrated as a second mechanism for robustification. This is particularly useful for temporary problems in the detection of ships. The extended Kalman filter (EKF) is used here, in which the state transition and the observation models are non-linear functions of the state and the input since the linearity of the ship movements (trajectory) is not guaranteed. For this purpose, the transition from $\mathbf{p}_i$ to $\mathbf{p}_{i+1}$ is modeled by an equation $f$ in the EKF, and state variables are updated in a prediction and correction cycle. The equations for the prediction step of the EKF are as follows~\cite{Mendes22}:
\begin{align}
\mathbf{x}_k  & = f(\mathbf{x}_{k-1})\\
\mathbf{P}_k & = \mathbf{A} \mathbf{P}_{k-1} \mathbf{A}^T + \mathbf{Q}\textrm{.}
\end{align}
$\mathbf{A}$ is the state transition matrix, which contains the properties of the physical model, and $\mathbf{Q}$ is the covariance matrix of the state transition, which models the noise. In the update step, the new value $\mathbf{x}_k$ is determined by calculating the Kalman gain $\mathbf{K}_k$ and taking measurements $\mathbf{z}_k$:
\begin{align}
\mathbf{K}_k  & = \mathbf{P}_k \mathbf{H}^T \left(\mathbf{H} \mathbf{P}_k \mathbf{H}  + \mathbf{R} \right)^{-1}\\
\mathbf{x}_k & = \mathbf{x}_k + \mathbf{K}_k(\mathbf{z}_k - h(\mathbf{x}_k))\\
\mathbf{P}_k & = \mathbf{P}_k - \mathbf{K}_k \mathbf{H} \mathbf{P}_k\textrm{.}
\end{align}
The EKF can, therefore, also enable tracking and prediction in cases where detection was temporarily disrupted (e.g., due to occlusions) and misinterpreted (e.g., due to artifacts).

\section{Experimental Results}
This section presents our experimental evaluation of different inland navigation and ocean settings with a varying number of vessels to detect, track, and predict. 

\subsection{Data sets}
Our data set for the evaluation consists of a total of nine videos. Five videos represent the vessel's perspective on inland waterways, especially canals, which are characterized by short-sight distances, mostly obstructed areas, and relatively close vessels. One video is recorded from a drone's perspective of the river Rhein. The other four videos cover situations on the North Sea, the Atlantic, and the Gulf of Thailand, showing only vessels or small islands.\footnote{Data sets are available at: \url{https://doi.org/10.5281/zenodo.14243467}}

\subsection{Analysis}
\begin{figure}[tbp]
\centering
\subfloat[Bounding boxes]{\includegraphics[width=.49\textwidth]{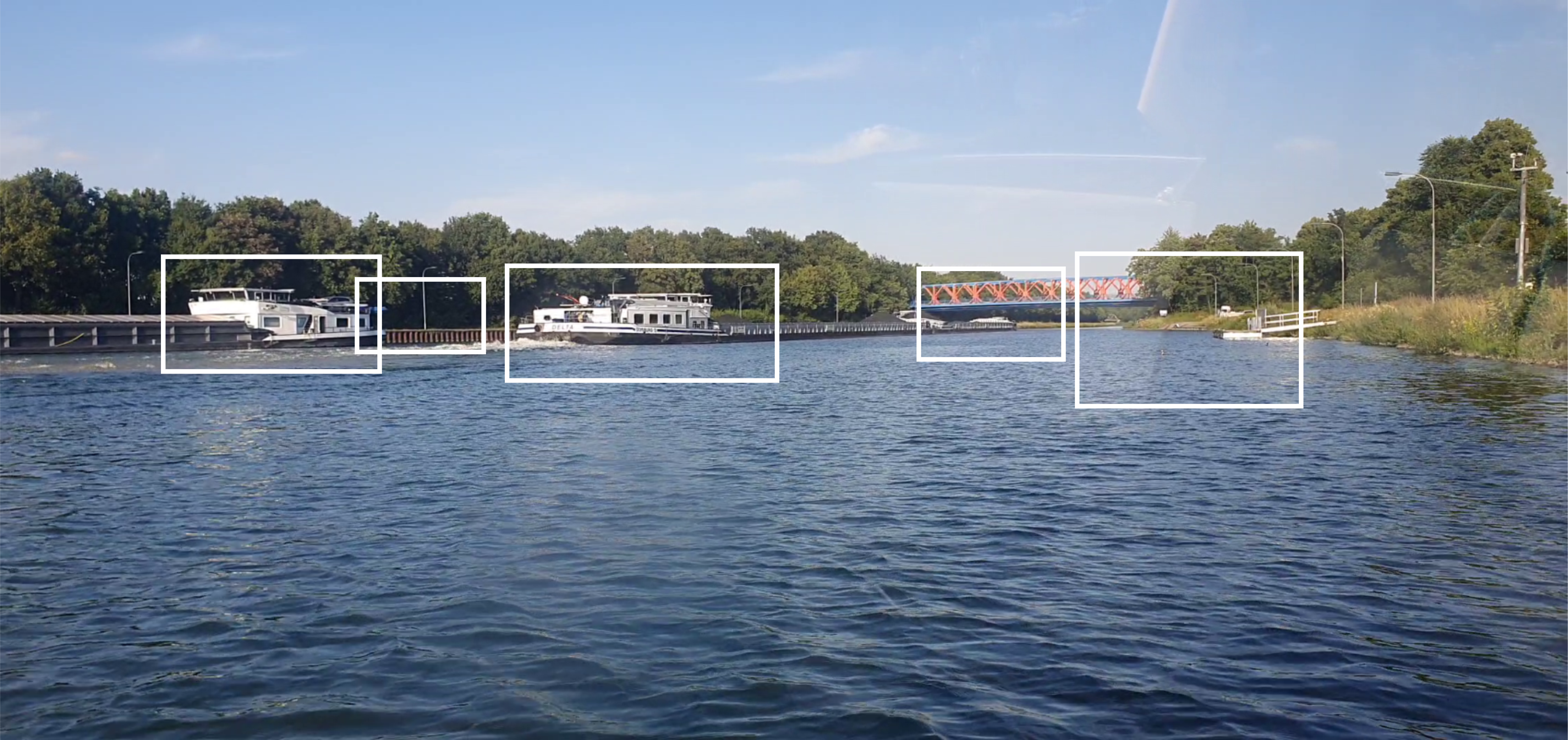}}\hfill
\subfloat[Class activation maps]{\includegraphics[width=.49\linewidth]{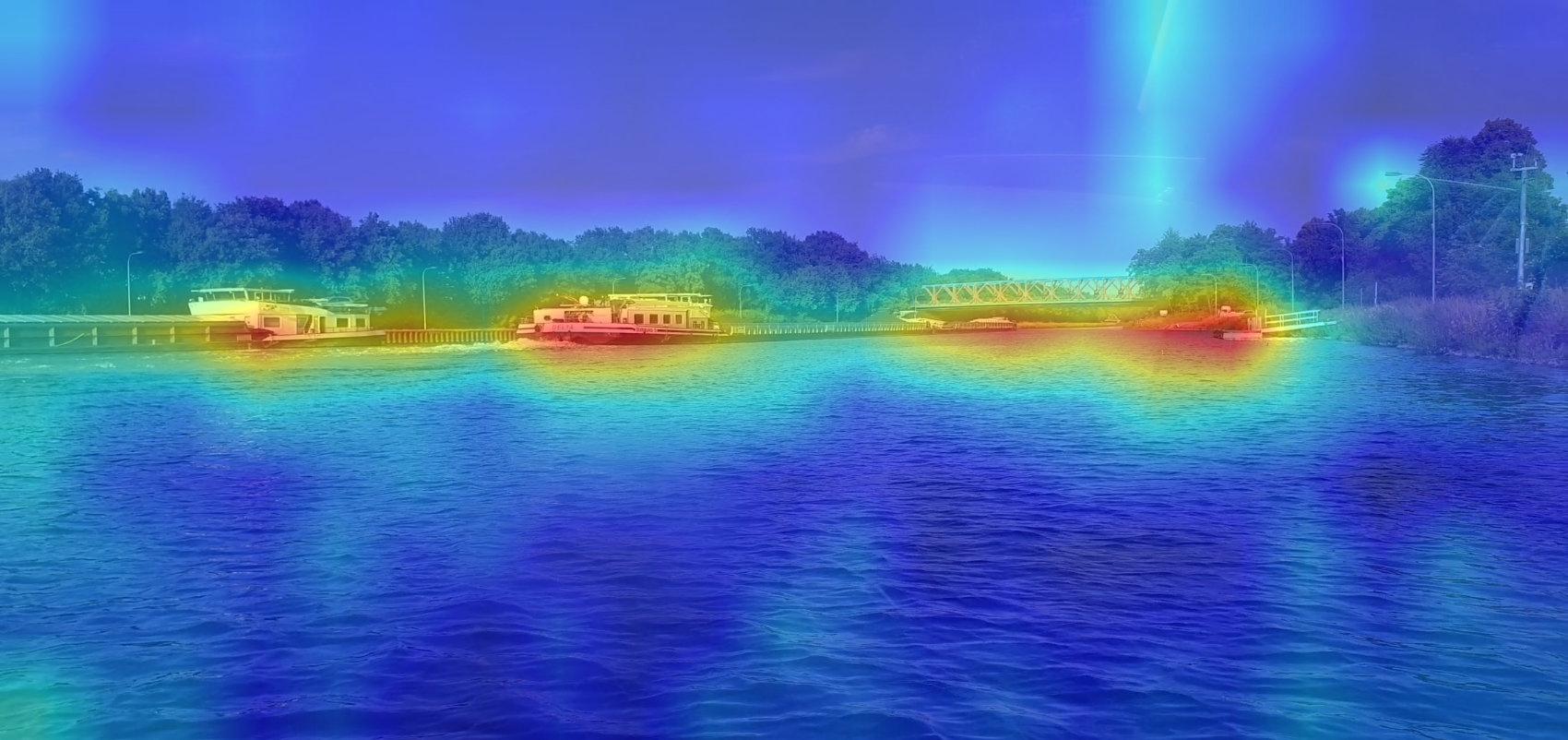}}		
	\caption{The class activation map presents a high confidence for many objects at the shoreline and a low confidence for the center section of the vessels. Furthermore, longer vessels are represented by multiple high-confidence values.}
\label{fig:detection_bb_cam_1}
\end{figure}

\begin{figure}[tbp]
\centering
\subfloat[Bounding boxes]{\includegraphics[width=.49\textwidth]{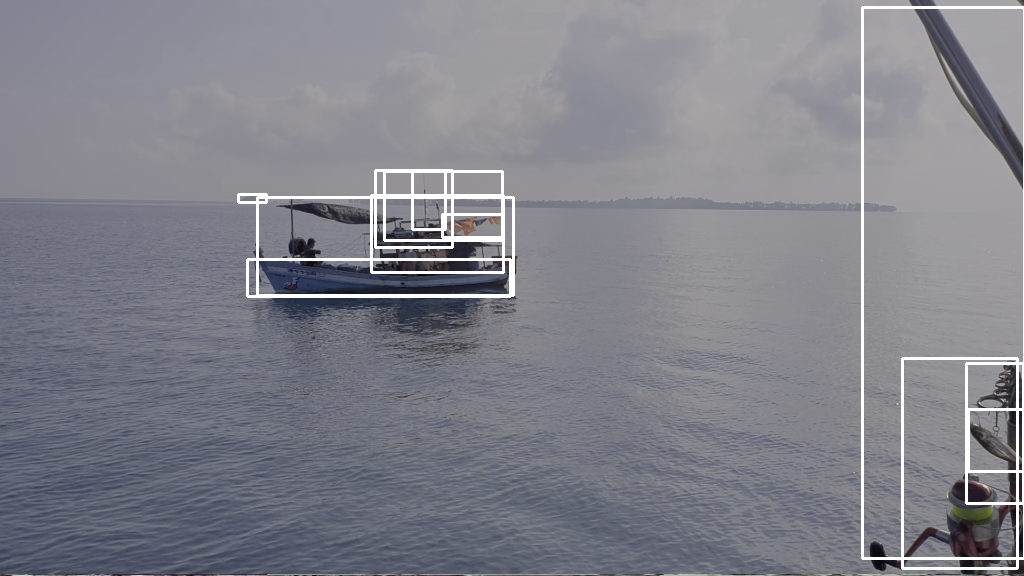}}\hfill
\subfloat[Class activation maps]{\includegraphics[width=.49\linewidth]{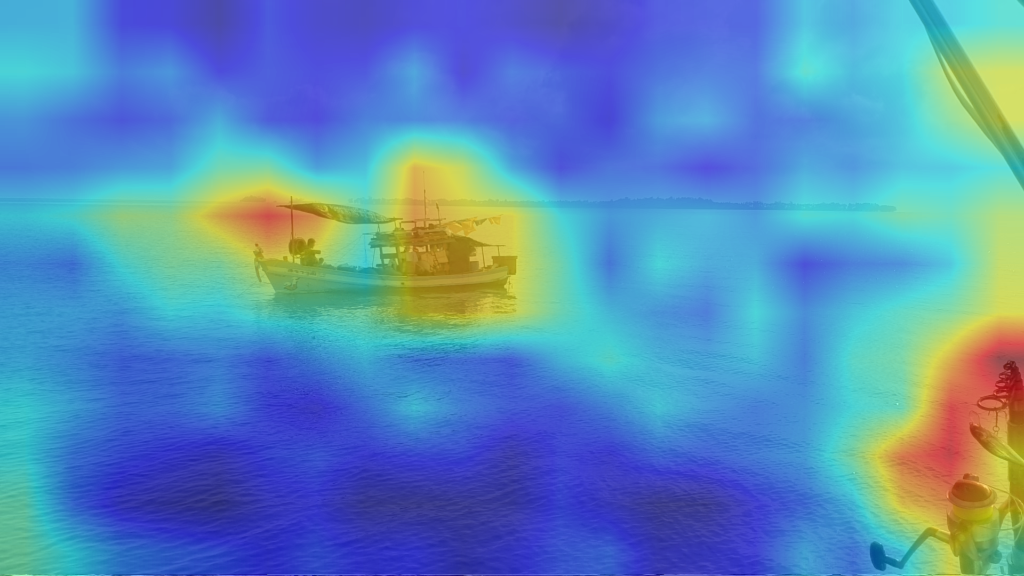}}
	\caption{The class activation maps present a high level of confidence for the fishing rod and the island in the background.}
\label{fig:detection_bb_cam_2}
\end{figure}
The standard Faster R-CNN model for detection, which is trained on the Microsoft Common Objects in Context~(COCO) data set~\cite{COCO}, shows a low precision for inland and sea-going vessel detection. However, the precision of inland vessels is slightly lower. CAM techniques are used to explain CNN detection. The CAM visualization highlights many objects on the water surface and at the shoreline with a higher to high confidence, which is visualized by yellow to red colors in Fig.~\ref{fig:detection_bb_cam_1}. Another example of sea-going vessels is presented in Fig.~\ref{fig:detection_bb_cam_2}, where a fishing rod and a small island in the background gain the highest confidence instead of a vessel. The increased recall compared to the precision and the F1 score indicate a large number of false positive detections, which proves our statement of a large number of wrongly classified detected objects as ships.

\begin{figure}[t]
    \centering
	\includegraphics[width=0.8\linewidth]{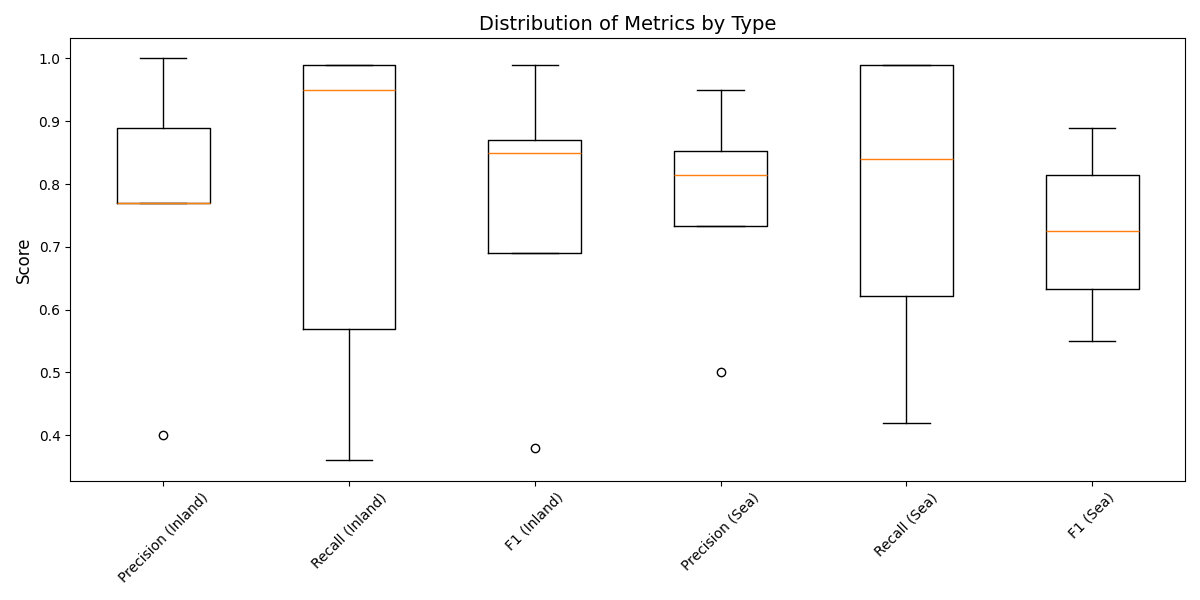}	
	\caption{Comparison of detection quality between inland and sea-going vessels}
\label{fig:detection}
\end{figure}

Figures~\ref{fig:detection_bb_cam_1} and \ref{fig:detection_bb_cam_2} show some of the challenges involved in detecting and tracking vessels in inland navigation: Objects in the foreground and background, several distinguishable ships simultaneously, longer, partly not completely visible vessels, and extensive details on the shoreline. This is also partly indicated in Fig.~\ref{fig:detection}. For inland navigation, the median is not the mean value of the central percentile.

\begin{figure}[t]
\centering
	\includegraphics[width=0.6\linewidth]{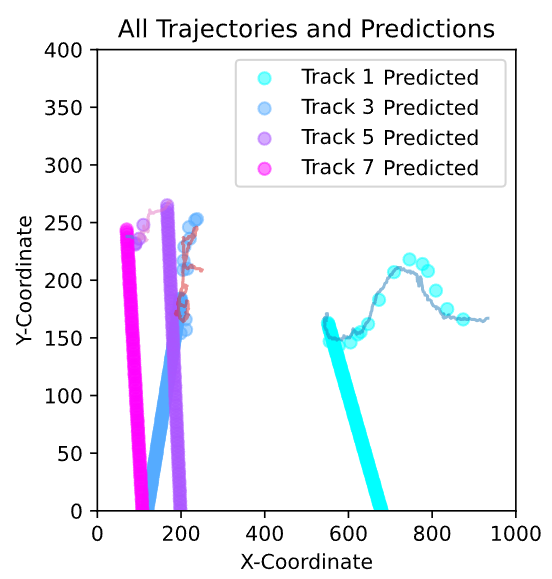}
	\caption{Comparison of predicted and real trajectory of vessels in Video~5. At first, the prediction assumes a linear movement, which later converges towards the real trajectory. For the underlying tracking, we used the ByeTrack tracker.}
\label{fig:prediction}
\end{figure}

\begin{figure}[tbp]
	\centering
	\includegraphics[width=0.9\linewidth]{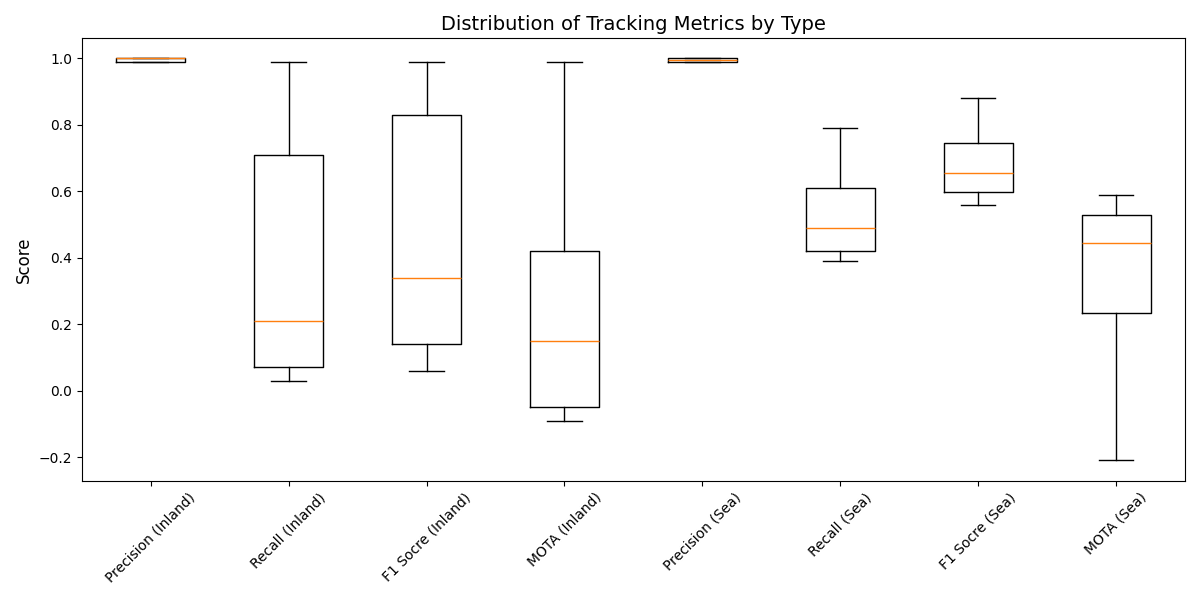}	
	\caption{Comparison of tracking quality between inland and sea-going vessels with BoT-SORT}
\label{fig:track_BoT-SORT}
\end{figure}

\begin{figure}[tbp]
	\centering
	\includegraphics[width=0.9\linewidth]{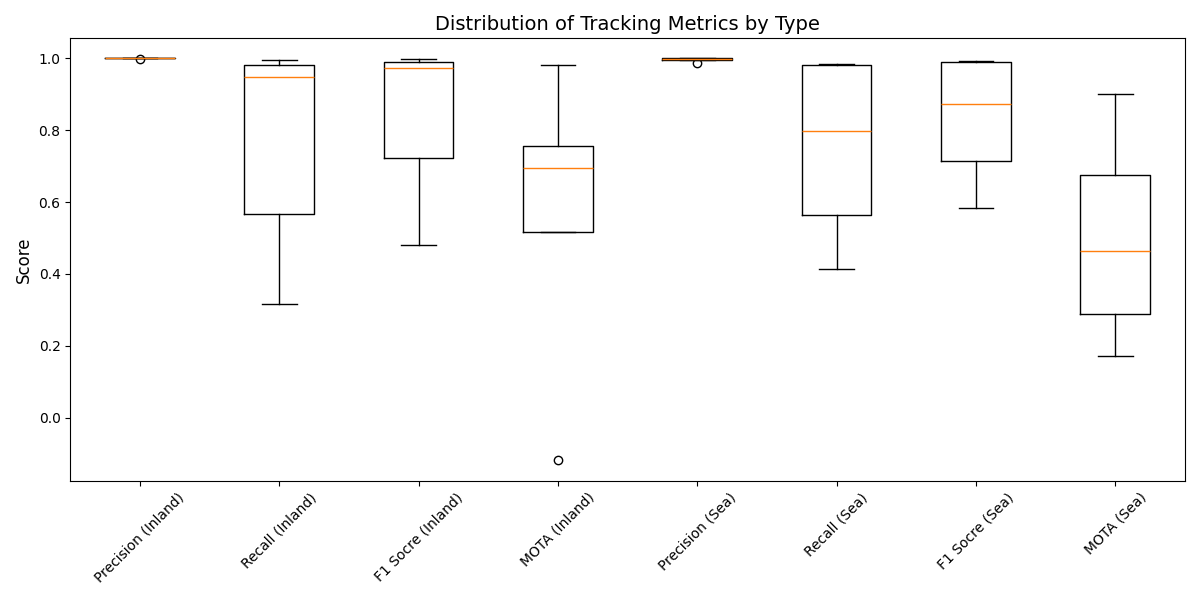}	
	\caption{Comparison of tracking quality between inland and sea-going vessels with Deep OC-SORT}
\label{fig:track_deep}
\end{figure}

\begin{figure}[tbp]
\centering
	\includegraphics[width=0.9\linewidth]{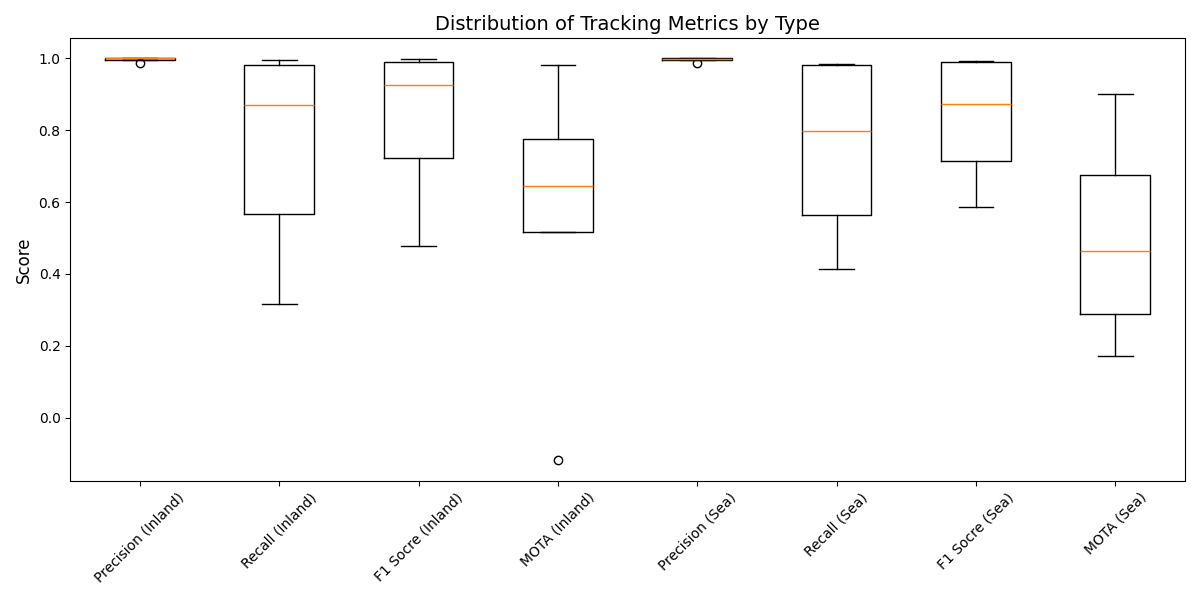}	
	\caption{Comparison of tracking quality between inland and sea-going vessels with ByeTrack}
\label{fig:track_byte}
\end{figure}

\begin{figure}[tbp]
	\centering
	\includegraphics[width=0.9\linewidth]{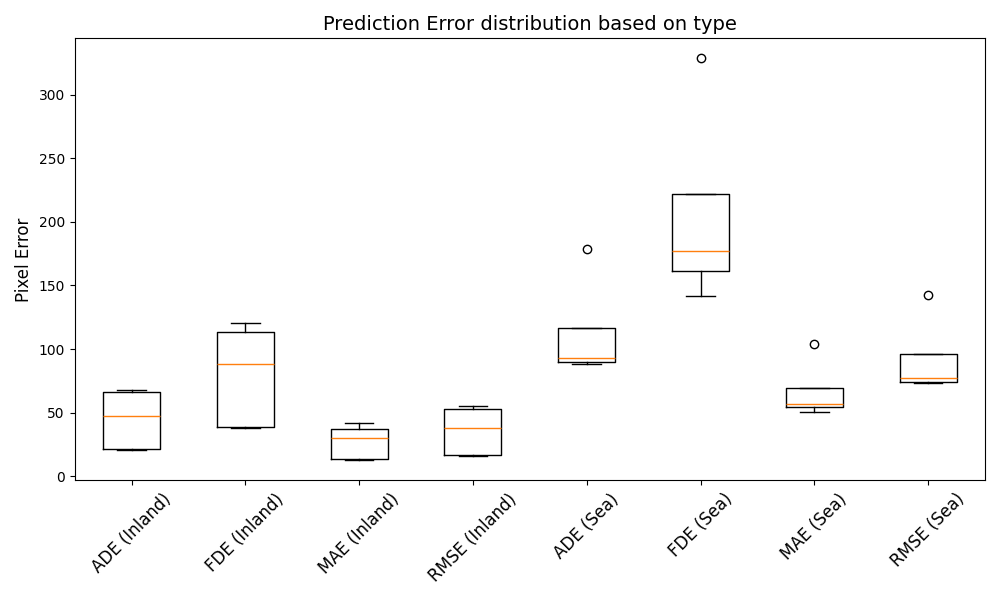}	
	\caption{Comparison of prediction quality between inland and sea-going vessels with BoT-SORT. We present the average distance error (ADE), the final distance error (FDE) of the prediction, the mean absolute error (MAE), and the root mean squared error (RMSE) for the categories of the video streams.}
\label{fig:predict_BoT-SORT}
\end{figure}

\begin{figure}[tbp]
	\centering
	\includegraphics[width=0.9\linewidth]{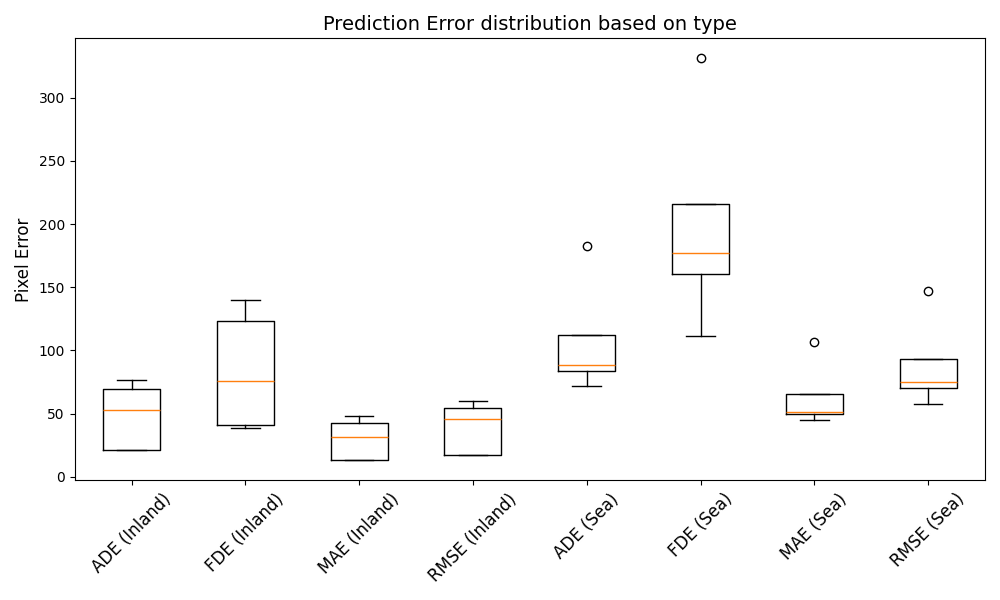}	
	\caption{Comparison of prediction quality between inland and sea-going vessels with Deep OC-SORT. We present the average distance error (ADE), the final distance error (FDE) of the prediction, the mean absolute error (MAE), and the root mean squared error (RMSE) for the categories of the video streams.}
\label{fig:predict_deep}
\end{figure}

\begin{figure}[tbp]
\centering
	\includegraphics[width=0.9\linewidth]{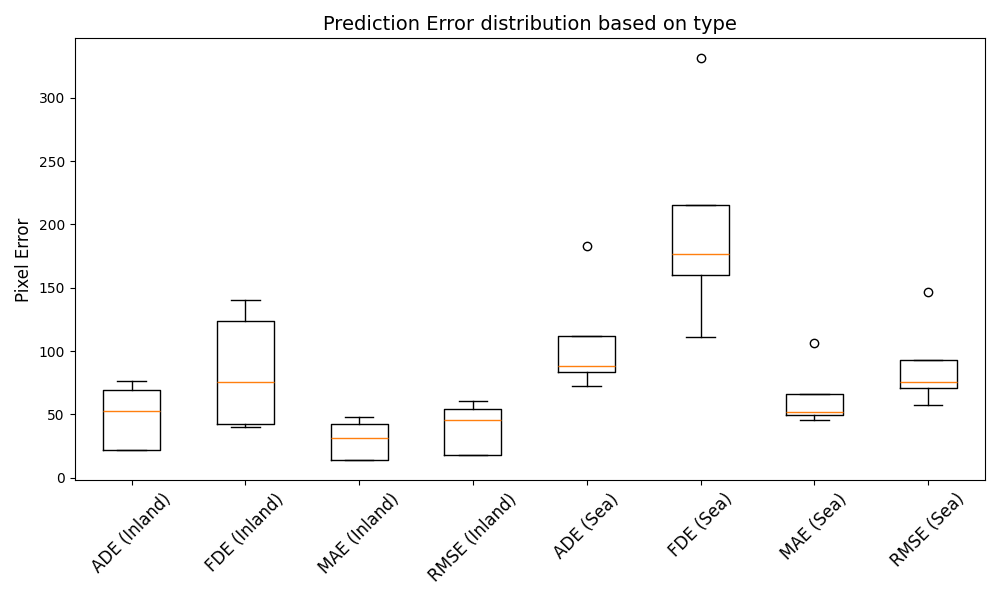}	
	\caption{Comparison of prediction quality between inland and sea-going vessels with ByeTrack. We present the average distance error (ADE), the final distance error (FDE) of the prediction, the mean absolute error (MAE), and the root mean squared error (RMSE) for the categories of the video streams.}
\label{fig:predict_byte}
\end{figure}

As shown in Fig.~\ref{fig:track_BoT-SORT}, \ref{fig:track_deep}, and \ref{fig:track_byte}, all three tracking algorithms have almost the same tracking precision. Consequently, the prediction as an implication of the tracking is almost identical, too (see Fig.~\ref{fig:predict_BoT-SORT}, \ref{fig:predict_deep}, and \ref{fig:predict_byte}). Major differences in prediction quality between inland and sea-going vessels can be observed. The four metrics, average distance error (ADE), final distance error (FDE), mean absolute error (MAE), and root mean squared error (RMSE), are expressed in pixel units. The video frames are resized to 1280 x 720, but the pixels cannot be correlated to metric distances due to different camera setups. In total, the error metrics are small and imply a useful prediction. Fig.~\ref{fig:prediction} shows a comparison between the predicted and real trajectories of vessels. The dots show the predicted positions of the vessels, and the curve shows the real trajectories. The curves show that the Kalman filter only leads to a deviating prediction in the short term (see, for example, the turquoise curve).



\subsection{Discussion}
The slow movements of vessels and the short amounts of overlapping situations lead to good tracking quality for almost all objects. But, it has to be considered that small ships, which are shadowed by larger vessels, are hidden for a long time period due to the slow movements. Hence, the tracking has to store the feature of each object for a longer duration to avoid id switching or tracking losses. This is one key difference to tracking systems for other vehicles.
The differences in prediction quality between inland and sea-going vessels can be based on the ships' movements. In contrast to the stabilized drone video of the Rhein river, the other videos are mainly captured on vessels which propagates its movement into the video.
The differences in wave height and overall conditions explain the reduced prediction quality.
Despite these differences, the prediction error metrics imply possible usages.

The main drawbacks of the detection algorithms are, that they are not optimized for vessel detection especially inland vessels.
In general, all non-extremely large objects on water surfaces like rivers, lakes, or open seas are classified as vessels. This includes fixed structures like platoons or docking ports, as shown in Fig.~\ref{fig:detection_bb_cam_1}. Thus, self-trained networks are essential for autonomous vessels and assistant systems for remote captains. 
The object tracking based on Kalman filters offers stable and smoothed trajectories even after a small number of frames with missing detection.
Due to the missing information of the camera setup, calibration and movement, the trajectories describe relative movements which are useful for remote captains with fixed installed cameras on the vessels or traffic control centers. To use the information and prediction in autonomous systems, e.g., for collision avoidance, a transformation based on the own odometry is crucial. In addition, it necessary for proving the correctness of AIS data.

\subsection{Conclusion}
We presented a methodology to combine existing systems to track and predict vessels on inland waterways and on the open sea. The numerical results indicate the functionality of our approach for tracking and prediction. However, the approach is mainly influenced by the low detection precision. Thereby, it was revealed that the state-of-the-art existing detection networks that are trained on the Microsoft Common Objects in Context~(COCO) data set~\cite{COCO} have to be adjusted, trained, and fine-tuned, especially for inland waterways, because of a high number of false positive detected vessels.
The proposed methodology and its implementation can be combined with a calibrated camera setup, enabling precise mapping of vessel positions to global coordinates as mandatory information for autonomous vessels. Thus, we will extend our inland vessel data sets to improve the detection rates and add a vessel classification to highlight potential risks of different ship types to remote captains. This classification is also required to identify applicable collision avoidance rules, which is impossible based on pure AIS, radar, and laser scanner data.

\bibliographystyle{plain}
\bibliography{Ship-Technology-Research}

\begin{thebibliography}{10}

\bibitem{aharon2022bot}
Nir Aharon, Roy Orfaig, and Ben-Zion Bobrovsky.
\newblock Bot-sort: Robust associations multi-pedestrian tracking.
\newblock {\em arXiv preprint arXiv:2206.14651}, 2022.

\bibitem{bo2021ship}
LI~Bo, XIE Xiaoyang, WEI Xingxing, and TANG Wenting.
\newblock Ship detection and classification from optical remote sensing images:
  A survey.
\newblock {\em Chinese Journal of Aeronautics}, 34(3):145--163, 2021.

\bibitem{carion2020end}
Nicolas Carion, Francisco Massa, Gabriel Synnaeve, Nicolas Usunier, Alexander
  Kirillov, and Sergey Zagoruyko.
\newblock End-to-end object detection with transformers.
\newblock In {\em European conference on computer vision}, pages 213--229.
  Springer, 2020.

\bibitem{DETR}
Nicolas Carion, Francisco Massa, Gabriel Synnaeve, Nicolas Usunier, Alexander
  Kirillov, and Sergey Zagoruyko.
\newblock End-to-end object detection with transformers.
\newblock In {\em European conference on computer vision}, pages 213--229.
  Springer, 2020.

\bibitem{ciaparrone2020deep}
Gioele Ciaparrone, Francisco~Luque S{\'a}nchez, Siham Tabik, Luigi Troiano,
  Roberto Tagliaferri, and Francisco Herrera.
\newblock Deep learning in video multi-object tracking: A survey.
\newblock {\em Neurocomputing}, 381:61--88, 2020.

\bibitem{diwan2023object}
Tausif Diwan, G~Anirudh, and Jitendra~V Tembhurne.
\newblock Object detection using yolo: Challenges, architectural successors,
  datasets and applications.
\newblock {\em multimedia Tools and Applications}, 82(6):9243--9275, 2023.

\bibitem{er2023ship}
Meng~Joo Er, Yani Zhang, Jie Chen, and Wenxiao Gao.
\newblock Ship detection with deep learning: a survey.
\newblock {\em Artificial Intelligence Review}, 56(10):11825--11865, 2023.

\bibitem{farhadi2018yolov3}
Ali Farhadi and Joseph Redmon.
\newblock Yolov3: An incremental improvement.
\newblock In {\em Computer vision and pattern recognition}, volume 1804, pages
  1--6. Springer Berlin/Heidelberg, Germany, 2018.

\bibitem{franke1991scattered}
Richard Franke and Gregory~M Nielson.
\newblock Scattered data interpolation and applications: A tutorial and survey.
\newblock {\em Geometric Modeling: Methods and Applications}, pages 131--160,
  1991.

\bibitem{hassan2024multi}
Saif Hassan, Ghulam Mujtaba, Asif Rajput, and Noureen Fatima.
\newblock Multi-object tracking: a systematic literature review.
\newblock {\em Multimedia Tools and Applications}, 83(14):43439--43492, 2024.

\bibitem{FastReID}
Lingxiao He, Xingyu Liao, Wu~Liu, Xinchen Liu, Peng Cheng, and Tao Mei.
\newblock Fastreid: A pytorch toolbox for general instance re-identification.
\newblock In {\em Proceedings of the 31st ACM International Conference on
  Multimedia}, MM '23, page 9664–9667, New York, NY, USA, 2023. Association
  for Computing Machinery.

\bibitem{he2022survey}
Mingwei He, Bohan Li, and Songlin Sun.
\newblock A survey of class activation mapping for the interpretability of
  convolution neural networks.
\newblock In {\em International Conference On Signal And Information
  Processing, Networking And Computers}, pages 399--407. Springer, 2022.

\bibitem{Henn.2018}
Rupert Henn and Berthold Holtmann.
\newblock Autonomes fahren in der binnenschifffahrt: Machbarkeitsstudie f{\"u}r
  ein testfeld im ruhrgebiet.

\bibitem{jie2021ship}
Yang Jie, LilianAsimwe Leonidas, Farhan Mumtaz, and Munsif Ali.
\newblock Ship detection and tracking in inland waterways using improved yolov3
  and deep sort.
\newblock {\em Symmetry}, 13(2):308, 2021.

\bibitem{khan2005survey}
Arshad Khan, Islam Khan, and Tariq Aziz.
\newblock A survey on parametric spline function approximation.
\newblock {\em Applied mathematics and computation}, 171(2):983--1003, 2005.

\bibitem{kohler2023few}
Mona K{\"o}hler, Markus Eisenbach, and Horst-Michael Gross.
\newblock Few-shot object detection: A comprehensive survey.
\newblock {\em IEEE Transactions on Neural Networks and Learning Systems},
  2023.

\bibitem{COCO}
Tsung-Yi Lin, Michael Maire, Serge Belongie, James Hays, Pietro Perona, Deva
  Ramanan, Piotr Doll{\'a}r, and C.~Lawrence Zitnick.
\newblock Microsoft coco: Common objects in context.
\newblock In David Fleet, Tomas Pajdla, Bernt Schiele, and Tinne Tuytelaars,
  editors, {\em Computer Vision -- ECCV 2014}, pages 740--755, Cham, 2014.
  Springer International Publishing.

\bibitem{liu2020deep}
Li~Liu, Wanli Ouyang, Xiaogang Wang, Paul Fieguth, Jie Chen, Xinwang Liu, and
  Matti Pietik{\"a}inen.
\newblock Deep learning for generic object detection: A survey.
\newblock {\em International journal of computer vision}, 128:261--318, 2020.

\bibitem{Deep_OC-SORT}
Gerard Maggiolino, Adnan Ahmad, Jinkun Cao, and Kris Kitani.
\newblock Deep oc-sort: Multi-pedestrian tracking by adaptive
  re-identification.
\newblock {\em arXiv preprint arXiv:2302.11813}, 2023.

\bibitem{Martelli.2021}
Michele Martelli, Antonio Virdis, Alberto Gotta, Pietro Cassara, and Maria {Di
  Summa}.
\newblock An outlook on the future marine traffic management system for
  autonomous ships.
\newblock {\em IEEE Access}, 9:157316--157328, 2021.

\bibitem{Mendes22}
Alexandre~Schalch Mendes, Pablo~Siqueira Meirelles, Janito~Vaqueiro Ferreira,
  and Eduardo~Rodrigues de~Lima.
\newblock Extended kalman filter for a monitoring system of the guyed towers.
\newblock In {\em Kalman Filter}, chapter~2. IntechOpen, 2022.

\bibitem{occorsio2021filtered}
Donatella Occorsio and Woula Themistoclakis.
\newblock On the filtered polynomial interpolation at chebyshev nodes.
\newblock {\em Applied Numerical Mathematics}, 166:272--287, 2021.

\bibitem{FasterRCNN}
Shaoqing Ren, Kaiming He, Ross Girshick, and Jian Sun.
\newblock Faster r-cnn: towards real-time object detection with region proposal
  networks.
\newblock In {\em Proceedings of the 28th International Conference on Neural
  Information Processing Systems - Volume 1}, NIPS'15, page 91–99, Cambridge,
  MA, USA, 2015. MIT Press.

\bibitem{sasikala2019ship}
J~Sasikala et~al.
\newblock Ship detection and recognition for offshore and inshore applications:
  a survey.
\newblock {\em International Journal of Intelligent Unmanned Systems},
  7(4):177--188, 2019.

\bibitem{shao2022deep}
Feifei Shao, Long Chen, Jian Shao, Wei Ji, Shaoning Xiao, Lu~Ye, Yueting
  Zhuang, and Jun Xiao.
\newblock Deep learning for weakly-supervised object detection and
  localization: A survey.
\newblock {\em Neurocomputing}, 496:192--207, 2022.

\bibitem{tracking2019}
Zahra Soleimanitaleb, Mohammad~Ali Keyvanrad, and Ali Jafari.
\newblock Object tracking methods:a review.
\newblock In {\em 2019 9th International Conference on Computer and Knowledge
  Engineering (ICCKE)}, pages 282--288, 2019.

\bibitem{YOLO_family}
Juan Terven, Diana-Margarita Córdova-Esparza, and Julio-Alejandro
  Romero-González.
\newblock A comprehensive review of yolo architectures in computer vision: From
  yolov1 to yolov8 and yolo-nas.
\newblock {\em Machine Learning and Knowledge Extraction}, 5(4):1680--1716,
  2023.

\bibitem{Theotokatos.2023}
Gerasimos Theotokatos, Joao Lucas~Dozzi Dantas, Georgia Polychronidi, Georgia
  Rentifi, and Marco~Molica Colella.
\newblock Autonomous shipping --- an analysis of the maritime stakeholder
  perspectives.
\newblock {\em WMU Journal of Maritime Affairs}, 22(1):5--35, 2023.

\bibitem{AttentionIsAll}
Ashish Vaswani, Noam Shazeer, Niki Parmar, Jakob Uszkoreit, Llion Jones,
  Aidan~N Gomez, \L~ukasz Kaiser, and Illia Polosukhin.
\newblock Attention is all you need.
\newblock In I.~Guyon, U.~Von Luxburg, S.~Bengio, H.~Wallach, R.~Fergus,
  S.~Vishwanathan, and R.~Garnett, editors, {\em Advances in Neural Information
  Processing Systems}, volume~30. Curran Associates, Inc., 2017.

\bibitem{wagner2010path}
Petr Wagner, Jiri Kotzian, Jan Kordas, and Viktor Michna.
\newblock Path planning and tracking for robots based on cubic hermite splines
  in real-time.
\newblock In {\em 2010 IEEE 15th Conference on Emerging Technologies \& Factory
  Automation (ETFA 2010)}, pages 1--8. IEEE, 2010.

\bibitem{ScoreCAM}
Haofan Wang, Zifan Wang, Mengnan Du, Fan Yang, Zijian Zhang, Sirui Ding, Piotr
  Mardziel, and Xia Hu.
\newblock Score-cam: Score-weighted visual explanations for convolutional
  neural networks.
\newblock In {\em Proceedings of the IEEE/CVF conference on computer vision and
  pattern recognition workshops}, pages 24--25, 2020.

\bibitem{wang2023semi}
Yanyang Wang, Zhaoxiang Liu, and Shiguo Lian.
\newblock Semi-supervised object detection: A survey on recent research and
  progress.
\newblock {\em arXiv preprint arXiv:2306.14106}, 2023.

\bibitem{wang2021machine}
Yu~Wang, G~Rajesh, X~Mercilin~Raajini, N~Kritika, A~Kavinkumar, and Syed
  Bilal~Hussain Shah.
\newblock Machine learning-based ship detection and tracking using satellite
  images for maritime surveillance.
\newblock {\em Journal of Ambient Intelligence and Smart Environments},
  13(5):361--371, 2021.

\bibitem{wang2022review}
Zhangu Wang, Jun Zhan, Chunguang Duan, Xin Guan, Pingping Lu, and Kai Yang.
\newblock A review of vehicle detection techniques for intelligent vehicles.
\newblock {\em IEEE Transactions on Neural Networks and Learning Systems},
  34(8):3811--3831, 2022.

\bibitem{Wojke2017}
Nicolai Wojke, Alex Bewley, and Dietrich Paulus.
\newblock Simple online and realtime tracking with a deep association metric.
\newblock In {\em 2017 IEEE International Conference on Image Processing
  (ICIP)}, pages 3645--3649, 2017.

\bibitem{Xu.2023}
Haitong Xu, L{\'u}cia Moreira, and C.~{Guedes Soares}.
\newblock Maritime autonomous vessels.
\newblock {\em Journal of Marine Science and Engineering}, 11(1), 2023.

\bibitem{zaidi2022survey}
Syed Sahil~Abbas Zaidi, Mohammad~Samar Ansari, Asra Aslam, Nadia Kanwal,
  Mamoona Asghar, and Brian Lee.
\newblock A survey of modern deep learning based object detection models.
\newblock {\em Digital Signal Processing}, 126:103514, 2022.

\bibitem{zhang2024development}
Chi Zhang, Xi~Zhang, Gui Gao, Haitao Lang, Genwang Liu, Chenghui Cao, Yuying
  Song, Yanan Guan, and Yongshou Dai.
\newblock Development and application of ship detection and classification
  datasets: A review.
\newblock {\em IEEE Geoscience and Remote Sensing Magazine}, 2024.

\bibitem{zhang2022bytetrack}
Yifu Zhang, Peize Sun, Yi~Jiang, Dongdong Yu, Fucheng Weng, Zehuan Yuan, Ping
  Luo, Wenyu Liu, and Xinggang Wang.
\newblock Bytetrack: Multi-object tracking by associating every detection box.
\newblock In {\em European conference on computer vision}, pages 1--21.
  Springer, 2022.

\bibitem{RT-DETR}
Yian Zhao, Wenyu Lv, Shangliang Xu, Jinman Wei, Guanzhong Wang, Qingqing Dang,
  Yi~Liu, and Jie Chen.
\newblock Detrs beat yolos on real-time object detection, 2023.

\bibitem{zhao2019object}
Zhong-Qiu Zhao, Peng Zheng, Shou-tao Xu, and Xindong Wu.
\newblock Object detection with deep learning: A review.
\newblock {\em IEEE transactions on neural networks and learning systems},
  30(11):3212--3232, 2019.

\bibitem{zhu2022traffic}
Yanzhao Zhu and Wei~Qi Yan.
\newblock Traffic sign recognition based on deep learning.
\newblock {\em Multimedia Tools and Applications}, 81(13):17779--17791, 2022.

\end{thebibliography}

\end{document}